\documentclass[letterpaper, 10 pt, conference]{ieeeconf}  

\IEEEoverridecommandlockouts                              

\usepackage{amsmath} 
\usepackage{multicol} 
\usepackage{multirow} 
\usepackage{relsize}
\usepackage[caption=false]{subfig}
\usepackage{cite}
\usepackage{graphicx}
\usepackage{hhline}
\usepackage{algorithm}
\usepackage[noend]{algpseudocode}
\usepackage{float}
\usepackage[table]{xcolor}
\usepackage{hyperref}

\algnewcommand{\IfThenElse}[3]{
  \State \algorithmicif\ #1\ \algorithmicthen\ #2\ \algorithmicelse\ #3}

\title{\LARGE \bf
A Peg-in-hole Task Strategy for Holes in Concrete
}

\author{Andr\'e Yuji Yasutomi$^{1,2}$, Hiroki Mori$^{2}$ and Tetsuya Ogata$^{2,3}$
\thanks{$^{1}$Andr\'e Yuji Yasutomi is with the Robotics Research Department, Center for Technology Innovation, R\&D Group, Hitachi, Ltd.
{\tt\footnotesize andre.yasutomi.ss@hitachi.com}} %
\thanks{$^{2}$Andr\'e Yuji Yasutomi, Hiroki Mori and Tetsuya Ogata are with the Department of Intermedia Art and Science, Graduate School of Fundamental Science and Engineering, Waseda University.
{\tt\footnotesize ogata@waseda.jp}}%
\thanks{$^{3}$Tetsuya Ogata is with the Artificial Intelligence Research Center, AIST.}
\thanks{Digital Object Identifier (DOI): \href{https://doi.org/10.1109/ICRA48506.2021.9561370}{10.1109/ICRA48506.2021.9561370}}
}

\begin{document}

\pubid{
\begin{tabular}[t]{@{}c@{}}~\copyright~2021 IEEE. Personal use of this material is permitted. Permission from IEEE must be obtained for all other uses, \\
                                    in any current or future media, including reprinting/republishing this material for advertising or promotional purposes, creating new collective works, \\ for resale or redistribution to servers or lists, or reuse of any copyrighted component of this work in other work\end{tabular}}

\maketitle

\begin{abstract}

A method that enables an industrial robot to accomplish the peg-in-hole task for holes in concrete is proposed. The proposed method involves slightly detaching the peg from the wall, when moving between search positions, to avoid the negative influence of the concrete's high friction coefficient. It uses a deep neural network (DNN), trained via reinforcement learning, to effectively find holes with variable shape and surface finish (due to the brittle nature of concrete) without analytical modeling or control parameter tuning. The method uses displacement of the peg toward the wall surface, in addition to force and torque, as one of the inputs of the DNN. Since the displacement increases as the peg gets closer to the hole (due to the chamfered shape of holes in concrete), it is a useful parameter for inputting in the DNN. The proposed method was evaluated by training the DNN on a hole 500 times and attempting to find 12 unknown holes. The results of the evaluation show the DNN enabled a robot to find the unknown holes with average success rate of 96.1$\%$ and average execution time of 12.5 seconds. Additional evaluations with random initial positions and a different type of peg demonstrate the trained DNN can generalize well to different conditions. Analyses of the influence of the peg displacement input showed the success rate of the DNN is increased by utilizing this parameter. These results validate the proposed method in terms of its effectiveness and applicability to the construction industry.

\end{abstract}
 
\section{INTRODUCTION}
Anchor-bolt insertion is a widely conducted task in the construction field. It involves inserting and hammering anchor bolts into holes pre-opened in concrete walls or floors to fix structural and non-structural elements to them \cite{anchor}. Since this task is tiresome and conducted in a dangerous, dirty environment, its automation is highly demanded \cite{construction1}. 

Anchor-bolt insertion is similar to the peg-in-hole task, which has been extensively studied in the robotics field. Like anchor-bolt insertion, the peg-in-hole task involves inserting an object, namely, the peg, into a hole with the same size and shape. However, in the case of the anchor-bolt insertion, the holes vary considerably in terms of surface finish and shape due to the brittle nature of the concrete. Moreover, the typical "press-and-slide" peg-in-hole strategy is difficult to apply since the high friction of the concrete causes: (i) torque overload on the robot joints, (ii) detachment of the anchor bolt from the end effector, and (iii) excessive noise in the measurements of force and moment. Those characteristics of concrete make it challenging to analytically model the interaction between peg and hole or to manually tune a control algorithm that can cope with all hole conditions for an effective task execution. In this paper, we propose a data-driven peg-in-hole method to solve the above challenges.

\subsection{Related work}
Proposed approaches to accomplish the peg-in-hole task involve analytical modeling \cite{peginhole_model1,peginhole_model2}, blind search \cite{peginhole_widesearch,peginhole_tiltandtouch}, visual servoing \cite{peginhole_visualservoing,peginhole_visualservoing2}, and learn-from-demonstration (LfD) \cite{peginhole_demonstration,peginhole_demonstration2,peginhole_demonstration4}. Analytical modeling involves deriving force and geometry models and using centering devices (i.e., remote center of compliances (RCC's)) to accomplish the peg-in-hole task. Blind search involves searching for the hole on the basis of predetermined trajectories and force/moment feedback. Visual servoing involves aligning the peg and hole with vision-based control. And LfD involves teaching robots to conduct tasks through multiple human demonstrations. Although these methods achieve high performance in real assembly settings, applying them to search for holes in concrete is challenging for the following reasons. First, analytical models, blind search, and LfD often rely on pressing and sliding the peg around the hole when searching for it, and that procedure is difficult on the concrete's high friction surface. Second, analytical models and LfD rely on the assumption that the shape and dimensions of the holes vary slightly, but that assumption is not the case for holes in concrete. Third, visual servoing and some LfD rely on visual feedback, which is not reliable in construction sites due to dust and constantly changing of light conditions.

\pubidadjcol

With the recent popularity of deep reinforcement learning (DRL) algorithms, some studies have adopted them to accomplish the peg-in-hole task with robot arms \cite{peginhole_rl1,peginhole_rl2,peginhole_rl3}. In those studies, peg-in-hole tasks were performed with precision that surpasses the positional repeatability of the robot arms used. Moreover, superior generalization capability in regard to different peg and hole sizes was demonstrated. However, in some studies, the absolute position of the peg has been used as one of the input parameters of a deep neural network (DNN) \cite{peginhole_rl1,peginhole_rl2}. This parameter might hinder the capability of the DNN to generalize to holes in different positions due to the dependency of the DNN on it. To avoid that problem, some studies use the relative position from a vision-estimated hole position \cite{peginhole_rl3}. Nevertheless, by using such a position parameter, the DNN might still learn to search for the hole with only a determined path that ignores force and moment data, and that outcome would negate the advantage of DNN-based methods compared with blind-search methods \cite{peginhole_widesearch}.

\begin{figure}[t]
\begin{center}
\includegraphics[width=\columnwidth]{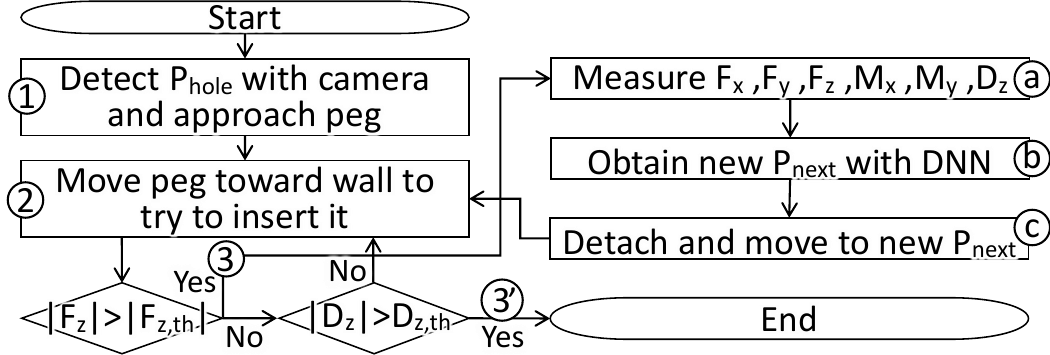}
\caption{Proposed method. $P_{hole}$ is the rough hole position obtained by, for example, a vision-based detection algorithm; $F_x$, $F_y$, and $F_z$ are the forces on X, Y, Z axes; $M_x$ and $M_y$ are the moments on the X and Y axes; $D_Z$ is peg displacement on the Z axis; and $P_{next}$ is the next search position obtained by the DNN. Subscript $th$ refers to threshold. }
\label{fig:method}
\end{center}
\end{figure}

\subsection{Contributions}
In this work, an off-policy, data-driven method that enables a industrial robot to effectively find holes in concrete to accomplish the peg-in-hole task is proposed. The main contributions of this work are: 
\begin{itemize}
\item A hole-search strategy that involves detaching the peg from the concrete surface between discrete hole-search positions to avoid problems related to the high friction coefficient of concrete;
\item The adoption of a DNN trained via reinforcement learning (RL) to find holes with variable surface finish;
\item Improved generalization capabilities of the DNN by introducing displacement of the peg (in addition to force and torque) as input for the DNN;
\item Capability of finding holes even without peg position as input so different hole positions can be generalized.
\end{itemize}

To the best of our knowledge, this is the first research that targets the performance improvement of peg-in-hole tasks in high-friction, brittle materials such as concrete. 
Evaluations of the method with an experimental setup similar to that of construction sites demonstrate the effectiveness of the method and its applicability to real construction sites.

\section{METHODS}\label{sec:methods}
\subsection{Proposed method}\label{subsec:method}

\begin{figure}[t]
\begin{center}
\includegraphics[width=\columnwidth]{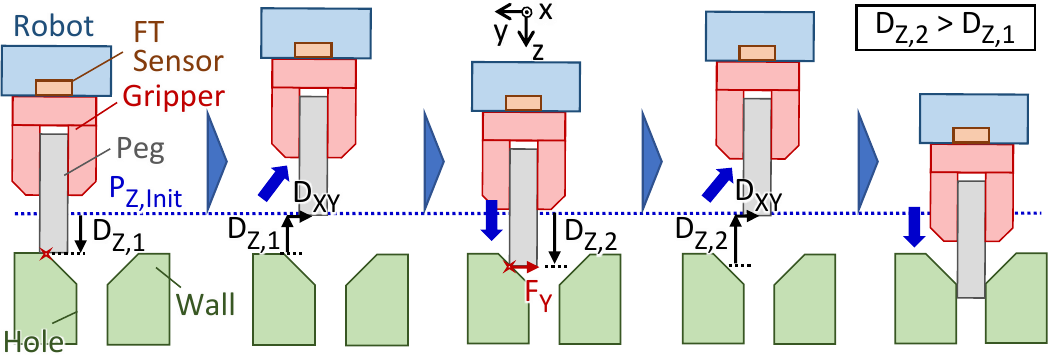}\\[-2ex]
\subfloat[\label{subfig:concept1} Step 1]{\hspace{.2\columnwidth}}
\subfloat[\label{subfig:concept2} Step 2]{\hspace{.2\columnwidth}}
\subfloat[\label{subfig:concept3} Step 3]{\hspace{.2\columnwidth}}
\subfloat[\label{subfig:concept4} Step 4]{\hspace{.2\columnwidth}}
\subfloat[\label{subfig:concept5} Step 5]{\hspace{.2\columnwidth}}
\caption{Usage example of proposed method. $P_{Z,init}$ is initial offset position from the wall, $D_Z$ is peg displacement from $P_{Z,init}$, and $D_{x,y}$ is peg displacement to the next search position in X (left/right) and Y (up/down).}
\label{fig:concept}
\end{center}
\end{figure}

The proposed method is shown in Fig. \ref{fig:method} and an example of its usage is illustrated in Fig. \ref{fig:concept}. The holes depicted in Fig. \ref{fig:concept} are chamfered in order to illustrate the real condition of holes opened in concrete (which inevitably become chamfer shaped due to the brittle nature of concrete). As shown in Fig. \ref{fig:method}, the method consists of: (1) approaching the hole on the basis of a rough estimation of the hole position (obtained by vision-based detection); (2) moving the peg toward the wall to tentatively insert it; (3) detaching the peg from the wall and moving it to the next position obtained by a DNN if the peg touches the wall, or (3') finishing the search if the hole is found.

The proposed method also involves using the displacement of the peg toward the wall, in addition to the force and moment, as input for the DNN. The motivation for using the displacement can be understood by comparing displacements $D_{z,1}$ and $D_{z,2}$ in Fig. \ref{fig:concept}. As shown in the figure, $D_{z,2}$ is greater than $D_{z,1}$ since in step 1 the peg contacts the wall outside the chamfer while in step 3 the peg contacts the chamfer region around the hole. This shows the displacement increases as the peg approaches the hole, which is a characteristic that makes the displacement a valuable parameter since it indicates the proximity to the hole.

Finally, the proposed method also involves using a DNN trained via RL to choose the next search position. A DNN was used instead of model-based algorithms since modeling the friction reactions of a brittle material such as concrete is challenging. Also, DNNs can generalize well to different system conditions \cite{dlbook}, which makes them suitable to cope with the variable surface finish of holes opened in concrete.

RL was used for training instead of a supervised training technique to eliminate the cumbersome dataset labeling process. Even if an algorithm was used to automatically label the dataset, acquiring the dataset while judging the actions with a reward seems to be less time consuming. Moreover, for the targeted task, a reward algorithm that ensured an effective training could be easily derived, and the exploring feature of the RL provides the DNN with more path options to find the hole under a given condition. 

The proposed method focus on the search phase of the peg-in-hole task. Thus, the following assumptions were made:
\begin{itemize}
\item The anchor bolt (also referred to as peg) is already grasped and positioned perpendicular to the wall;
\item Holes were opened perpendicular to the wall surface;
\item After the anchor was partially inserted in the hole, the anchor is hammered to completely insert it into the hole.
\end{itemize}
These assumptions are in line with real on-site applications since (i) wall orientation can be measured with, for example, a laser sensor, (ii) hole drilling can be automated and kept perpendicular to the wall, and (iii) we observed anchor hammering and the brittle nature of concrete significantly reduce anchor jamming caused by orientation misalignments of peg and hole.

\subsection{Deep reinforcement learning algorithm}\label{subsec:trainmethod}

\begin{figure}[t]
\begin{center}
\includegraphics[width=\columnwidth]{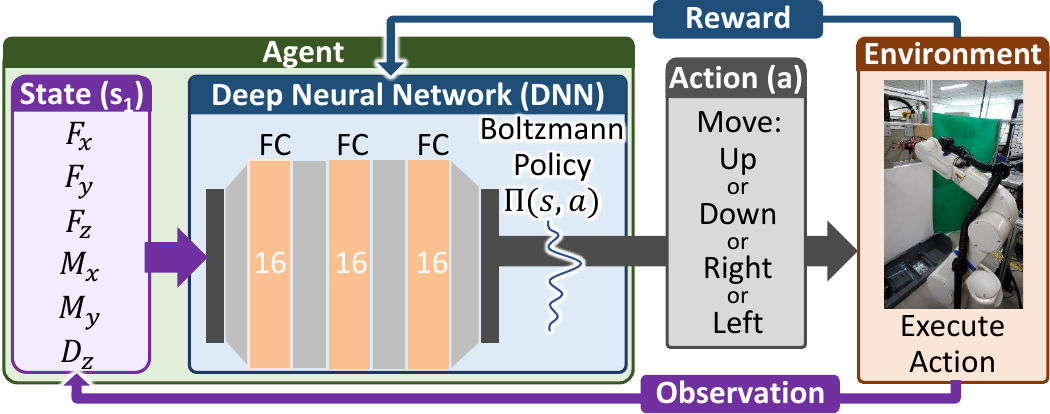}
\caption{Deep Q-Learning architecture}
\label{fig:nn}
\end{center}
\end{figure}

As the RL algorithm for training the DNN of the proposed method (hereinafter ``deep RL'' or DRL algorithm), \textit{deep q-learning} was selected \cite{doubleqlearning}. This algorithm was selected since its proved success in accomplishing tasks that are discretized into a limited number of actions was a good fit to the discretized search strategy adopted to avoid the concrete's friction \cite{doubleqlearning}. Also, the low dimensionality of the input data did not require a complex algorithm, with laborious parameter tuning, to be converted into task-effective actions. 

The \textit{deep q-learning} architecture used is shown in detail in Fig. \ref{fig:nn}. The agent of the architecture consists of a DNN and the environment state. The DNN used was the feed-forward DNN (hyper-parameters in Table \ref{tab:hyperparameters}) since this application does not have unobservable states; thus, it was assumed actions could be determined based on only the current state, not requiring recurrent neural networks. The state observed consists of the forces in the X, Y, and Z axes, the moments in the X and Y axes, and the robot displacement $D_{z}$  ($s_1 = [F_x,F_y,F_z,M_x,M_y,D_z]$)). A state that replaces $D_z$ with the moment in the Z axis $M_z$  ($s_2 = [F_x,F_y,F_z,M_x,M_y,M_z]$)) was also used in a separate training to analyze the effectiveness of using $D_z$ as the DNN input. The actions output by the DNN and executed by the robot in the environment were to move the anchor bolt up (+Y), down (-Y), to the right (-X) or to the left (+X) by a displacement $D_{xy}$. To update the weights of the \textit{deep q-learning} network (DQN), the equation below based on the Belmann equation was used\cite{qlearning_bellman}: 

\begin{equation}\label{eq:bellman}
\begin{aligned} 
  \theta \leftarrow & \theta + \alpha (r + \gamma \cdot  \max{Q_{target}(s',a')} - Q(s,a))\nabla Q(s,a)
\end{aligned}
\end{equation}
Here, $Q$ is the q value of the main network, $Q_{target}$ is the q value of the target network (a copy of the main network for estimation), $s$ is the current state of the robot, $s'$ is the next state, $a$ is the current action, $a'$ is the next action, $\alpha$ is the learning rate, $r$ is the reward, $\gamma$ is a discount factor, and $\nabla$ is the gradient function \cite{peginhole_rl1}. \textit{Boltzmann exploration} \cite{boltzmann} was used to enhance environment exploration, which is given by
\begin{align} \label{eq:boltzmann}
  P(a) = \frac{exp(Q_t(a)/\tau)}{\sum_{i=1}^{n} exp(Q_t(i)/\tau)},
\end{align}
where $\tau$ is Boltzmann parameter, and $n$ number of actions.

\begin{table}[t]
\begin{center} 
\caption{Hyper-parameters used for the DRL algorithm}
\label{tab:hyperparameters} 
\begin{tabular}{rc|rc} \hline
\bf{Number of hidden layers}			& 3 	& \bf{Optimizer}			& Adam	\\  
\bf{Neurons per hidden layer}		& 16		& \bf{Batch size} & 32    \\ 
\bf{Activation hidden layers}		& ReLU		& \bf{Learning rate  ($\alpha$)}		& 0.001	\\ 
\bf{Activation last layer}		& Linear		& \bf{Discount factor ($\gamma$)}	& 0.99	\\ 
\bf{Boltzmann parameter ($\tau$)} & 1		& \bf{Distance limit ($D$)}		& 4 mm	\\ 
\multicolumn{3}{r}{\bf{Episodes for target network update}}  & 100  \\  
\multicolumn{3}{r}{\bf{Reward when hole is found ($r_{foundhole}$)}}	& 100	\\ 
\multicolumn{3}{r}{\bf{Max. number of steps ($k_{max}$)}}		& 100	\\ 
\multicolumn{3}{r}{\bf{Force threshold on the Z axis ($F_{z,th}$)}}	& 20 N	\\ 
\multicolumn{3}{r}{\bf{Displacement threshold on the Z axis ($D_{z,th}$)}}	& 6 mm	\\
\multicolumn{3}{r}{\bf{Displacement in the XY plane ($D_{xy}$)}}	& 1 mm	\\ \hline
\end{tabular}
\end{center}
\end{table}

The reward for each step is $r=-1$ when the RL episode did not end, and it is given by the equation below otherwise:
\begin{align} \label{eq:reward}
  r &=
	\begin{cases}
		r_{found hole},& \resizebox{.17\columnwidth}{!}{$\textrm{if hole found,}$}\\
		0,& \resizebox{.4\columnwidth}{!}{$\text{if }  d \leq d_0 \text{ and hole not found,}$ } \\
		-r_{found hole}\cdot \frac{d-d_0}{D-d_0} ,& \resizebox{.4\columnwidth}{!}{$\text{if }  d > d_0 \text{ and hole not found,}$ }
	\end{cases}
\end{align}
Here, $r_{foundhole}$ is the reward when the hole is found, $d_0$ is the initial distance from the hole, $d$ is the final distance, and $D$ is the distance limit. The negative reward at the end of each step makes the DRL algorithm strive to minimize the number of steps. The final reward at the end of the episode makes the DRL algorithm strive to approximate the peg to the hole position in order to avoid the negative reward when $d$ is greater than $d_0$. Total reward $R$, which is the sum of all rewards and an indicative of the performance of the DRL algorithm in each episode, is less or equal to zero when the hole is not found, and greater than zero when it is found.

The episode was set to end in case (i) the robot took more than $k_{max}$ steps, (ii) the peg trespassed a boundary limit ($d>D$), or (iii) the peg was inserted into the hole. Insertion of the peg was identified as when the force along the Z axis $|F_z|$ was lower than $F_{z,th}$ and displacement $D_z$ was greater than $D_{z,th}$, as presented in Fig. \ref{fig:method}. The values used in the DRL algorithm are listed in Table \ref{tab:hyperparameters}. 

\textit{Experience replay} \cite{exp_replay}, a technique that involves updating the DNN with previous states, actions, and rewards stored in a buffer, was used with a maximum buffer size of 10,000 experiences to improve the convergence of the DNN.

\subsection{Displacement importance analysis}
To analyze the effectiveness of $D_z$ as an input parameter for the network, the evaluation results for input states $s_1$ and $s_2$ were compared, and \textit{Guided backpropagation} \cite{guidedbp} was used to create an input-importance map (hereinafter “saliency'' map) of the inputs. This state-of-the-art saliency mapping algorithm was used because \textit{SmoothGrad} \cite{smoothgrad}, \textit{Grad-CAM} \cite{gradcam}, and \textit{Integrated Gradients} \cite{integratedgrad} were considered unsuitable. This is because \textit{SmoothGrad} requires an input with high dimension for its effectiveness, \textit{Grad-CAM} is applied to convolutional neural networks, and \textit{Integrated Gradients} presented high saliency for input $F_z$, which is a parameter that varies minutely and, thus, should not generate high saliency values. \textit{Guided backpropagation} \cite{guidedbp}, however, showed the lowest saliency values for input $F_z$ and reasonable values for the other inputs. Thus, it was considered suitable to analyze $D_z$.

\subsection{Comparison with model-based approaches}
To compare the performance of the hole search with the DNN and with model-based methods, hole search was also conducted through a blind search method, based on \cite{peginhole_widesearch}, and a moment-feedback-based method. The blind search method involved moving the peg using a decided spiral trajectory with steps spaced 1 mm from each other. The moment-based search method involved moving the peg toward the maximum force direction, when the peg is inside the chamfer region, and toward the direction the peg tilts otherwise. The peg presence inside or outside the chamfer region was judged by comparing the peg displacement with an initial peg displacement measured when the peg touches outside the chamfer. To move between search positions, the peg was detached from the concrete's surface in the same way as by the proposed method.

\section{EXPERIMENTAL SETUP AND CONDITIONS}\label{sec:setupandconditions}
\subsection{Experimental Setup}\label{subsec:setup}
\begin{figure}[t]
\begin{center}
\includegraphics[width=\columnwidth]{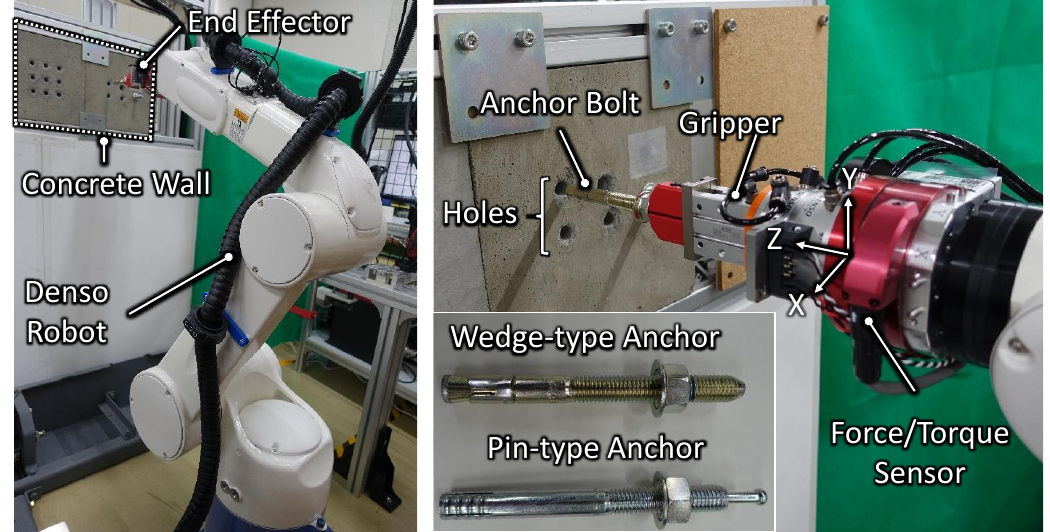}\\[-2ex]
\subfloat[\label{subfig:denso1} Whole system]{\hspace{.4\columnwidth}}
\subfloat[\label{subfig:denso2} End effector and anchor bolt details]{\hspace{0.6\columnwidth}}
\caption{Experimental setup for inserting anchor bolt}
\label{fig:denso}
\end{center}
\end{figure}

The setup used to train the DNN and validate the proposed method is shown in Fig. \ref{fig:denso}. A Denso robot (VM-60B1) was used to search for holes opened in a concrete wall to insert an anchor bolt into them. Two types of anchor bolt were available, but only the wedge-type anchor was used for training and main method evaluations. The holes are 12.7 mm in diameter, while the anchor bolts used are both 12 mm in diameter, so the clearance between the holes and the anchor bolts is 0.7 mm. To hold each anchor bolt during the search-and-insertion operation, an air gripper was used. The gripper was attached to the robot via a force-torque (FT) sensor, DynPick\textsuperscript{\tiny\textregistered} WEF-6A1000-30, fixed to the robot flange.

\begin{figure}[t]
\begin{center}
\includegraphics[width=\columnwidth]{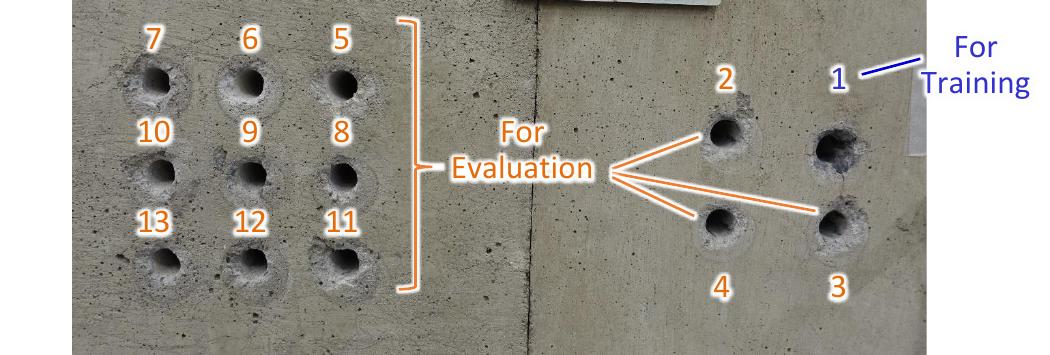}
\caption{Holes for training (in blue) and evaluations (in orange)}
\label{fig:holes}
\end{center}
\end{figure}

The holes used for training and evaluation are shown in detail in Fig. \ref{fig:holes}. They were opened into two different concrete blocks (forming a concrete wall) according to the conventional procedure used in construction. As shown in the figure, the holes become inevitably chamfered due to the brittle nature of the concrete. 

\begin{figure}[t]
\begin{center}
\includegraphics[width=\columnwidth]{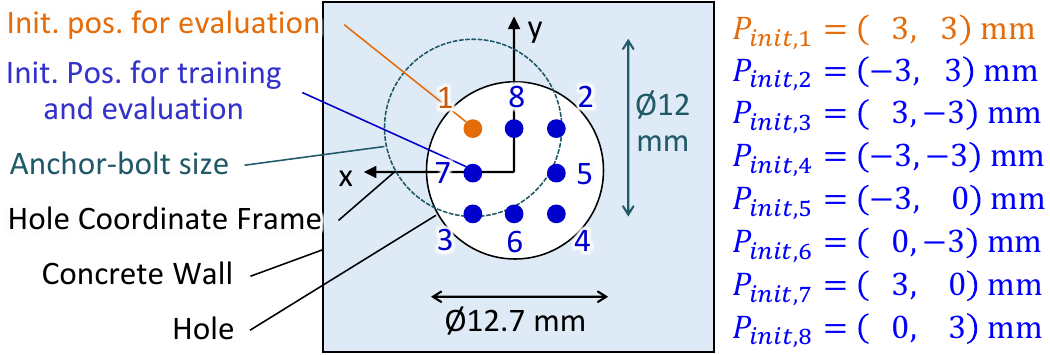}
\caption{Initial peg positions for training and evaluation}
\label{fig:traincond}
\end{center}
\end{figure}

The initial positions related to the hole center, which were used for training and evaluation are shown in Fig. \ref{fig:traincond}.

The initial positions were set 3 mm away from the hole origin, since that distance is the maximum visual-detection-and-positioning error estimated for the current setup.  Movement of the robot during training and evaluation was controlled according to the flow chart shown in Fig. \ref{fig:method}, where the forces and moments were measured by the FT sensor, and $D_z$ was calculated from the XYZ position of the tip of the anchor bolt obtained through forward kinematics. To compensate for the gravity, the FT sensor was zeroed right before the hole search. At the end of each episode, the robot was set to return to a home position and then move to the next initial position. The anchor was positioned perpendicularly to the wall by measuring the wall orientation with a laser sensor and making the robot flange parallel to it. 

\subsection{Training conditions}\label{subsec:traincond}
The DQN was trained on hole 1 by starting the peg from initial positions 2 to 8 ($P_{init,2}$ to $P_{init,8}$) chosen randomly at the start of each episode. Initial position 1 ($P_{init,1}$) was left to be used for preliminary evaluation (Fig. \ref{fig:traincond}). Both input states $s_1$ and $s_2$ were used to train separate networks to search for the hole. Each network was trained for 500 episodes.

Since training was conducted by repeatedly contacting the hole surroundings with an anchor bolt, the hole borders would wear down constantly, making them more chamfered every step. Since a hole overly chamfered due to long trainings does not reflect the real condition of the hole right after being opened, over-usage of hole 1 was avoided.

\subsection{Evaluation conditions}
The proposed method was evaluated first by attempting to find hole 1 (used for training) for 50 times, starting from the unknown initial position $P_{init,1}$. Once the method proved effective, it was evaluated by attempting to find the unknown holes (holes 2 to 13 in Fig. \ref{fig:holes}), starting from all the initial positions. For the unknown holes, hole search was attempted 25 times per initial position for holes 2 to 4, and 10 times per initial position for holes 5 to 13, making a total of 1320 attempts. Both state types $s_1$ and $s_2$ were used for the evaluations. 

To further assess the generalization capability of the proposed method, the method was evaluated with random peg initial positions chosen within 2 and 3 mm from the hole center on the X and Y axes, with the positions spaced 0.1 mm from each other. It was also evaluated with the pin-type anchor bolt, which is different from the one used for training. Both evaluations covered holes 8 to 11 for 100 episodes each.  

\section{RESULTS}\label{sec:results}

\subsection{Training results}\label{subsec:trainingres}

\begin{figure}[t]
\begin{center}
\includegraphics[width=\columnwidth]{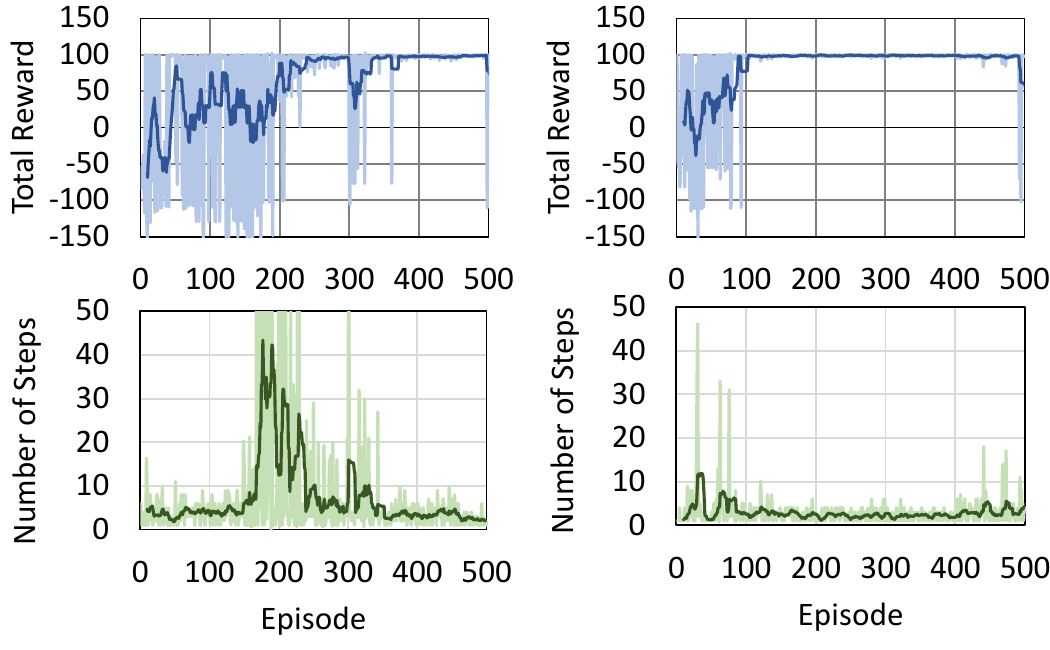}\\[-2ex]
\subfloat[\label{subfig:trainresultss1} $s_1$: With Dz]{\hspace{.6\columnwidth}}
\subfloat[\label{subfig:trainresultss2} $s_2$: Without Dz]{\hspace{0.4\columnwidth}}
\caption{Training results. Light color: raw data; dark color: moving average with 10 episodes.}
\label{fig:trainresults}
\end{center}
\end{figure}

The training results are presented in Fig. \ref{fig:trainresults}. As shown in Fig. \ref{subfig:trainresultss1}, the total reward for $s_1$ increases gradually and converges to a value near the maximum reward of 100 after about 350 episodes. This result shows the neural network failed to command the robot to insert the anchor bolt at first, as indicated by the negative rewards, but it gradually improved its performance the more it was trained. As for number of steps, it peaks at about 50 steps within episodes 150 and 250, but it decreases to values up to 10 steps after about 350 episodes of training. The peaks in number of steps suggest the DNN learned to avoid leaving the search boundaries, but it required more data to converge to hole discovery. These results suggest the neural network learned to insert the peg after 350 episodes. 

The results for input state $s_2$ are shown in Fig. \ref{subfig:trainresultss2}. For this input state, the charts indicate convergence was reached after 100 episodes which is earlier than for input state $s_1$. Moreover, the number of steps remained low during the whole training, barely passing 10 steps per episode. This result suggests the DNN learned to find the hole earlier than in the case of state $s_1$. However, it also suggests the DNN had less opportunity to explore  and learn from the environment which limits its generalization capability.

\subsection{Evaluation results}

The evaluation results for unknown initial peg position $P_{init,1}$ in hole 1 for input-state types $s_1$ and $s_2$ are listed in Table \ref{tab:evalres}. As indicated in the table, the proposed method effectively found hole 1 even for an unknown initial position (100$\%$ success rate) for both input-state types, in a relatively short time (below 7.8 seconds). 

The evaluation results for unknown holes 2 to 13 are also listed in Table \ref{tab:evalres}. As listed, the holes were correctly found for both input states with high success rates (up to 96.1$\%$) and low execution time (below 12.5 seconds). However, as expected, success rate was higher for evaluations with input state $s_1$, namely, 96.1$\%$ against 93.4$\%$. This result demonstrates peg displacement input $D_z$ improved the effectiveness of the DNN. Errors in the search with input state $s_2$ occurred mainly in the cases of holes 3 and 4, particularly for initial peg position $P_{init,1}$, which was the initial position that was not used for training in hole 1. This result suggests input state $s_2$ did not provide enough data to enable the network to learn to find the hole with this initial position. It is important to note the DNN trained with $s_1$ achieved a longer average execution time per episode due to the higher execution time required to find the holes in general. The execution time to find hole 3 and 5 with $s_2$ were exceptionally short because the anchor rapidly crossed the search boundaries, ending the episode in a few steps.

The evaluation results show the proposed method enables the robot to find holes independently of hole position, showing its superiority compared to DRL-based methods that depend on this parameter.

\begin{table}[t]
\begin{center} 
\caption{Evaluation results for unknown init. pos. ($P_{init,1}$ in hole 1) and unknown holes (holes 2 to 13). Numbers mentioned in the result analysis are highlighted in gray.}
\label{tab:evalres} 
\setlength\tabcolsep{2.5pt} 
\begin{tabular}{c|c|ccc|ccc}
\multirow{3}{*}{\bf{Hole}} &  \multirow{3}{*}{\shortstack{\bf{Init.}\\ \bf{Pos.} \\ \bf{No.}}} & \multicolumn{3}{c|}{\bf{State 1 ($s_1$)}} & \multicolumn{3}{c}{\bf{State 2 ($s_2$)}} \\ \cline{3-8}
&    & \multirow{2}{*}{\shortstack{\bf{Avg.}\\\bf{Time [s]}}} & \multirow{2}{*}{\shortstack{\bf{Avg.}\\\bf{Reward}}} & \multirow{2}{*}{\shortstack{\bf{Success}\\\bf{rate [$\%$]}}} & \multirow{2}{*}{\shortstack{\bf{Avg.}\\\bf{Time [s]}}} & \multirow{2}{*}{\shortstack{\bf{Avg.}\\\bf{Reward}}} & \multirow{2}{*}{\shortstack{\bf{Success}\\\bf{rate [$\%$]}}} \\ 
&   &  &  & & &  &       \\ \hline
1  			& $1$ & \cellcolor{gray!20}7.8 &	97.88 &	\cellcolor{gray!20}100 &	 \cellcolor{gray!20}7.2 &	98.12 &	\cellcolor{gray!20}100 \\   \hline\hline
2  	             & 1-8    & 6.0 &  93.66 &   97.5 &    5.5. &	98.80 &	100 \\ \hline
\multirow{3}{*}{3}  & 1 & \cellcolor{gray!20}19.7 &	93.12 &	100 &	6.9 &	-77.48 & \cellcolor{gray!20}8   \\ 
                    & 2-8 & 21.04 &91.15 &	98.6 &	7.62 &	88.9 &	94.9 \\ \cline{2-8}
                    & 1-8    & \cellcolor{gray!20}20.9 &	91.4 &	98.8 &	\cellcolor{gray!20}7.5 &	68.10 &	\cellcolor{gray!20}84.0 \\ \hline
\multirow{3}{*}{4}  & 1 & 8.5 &	97.60 &	100 &	12.5 &	43.97 &	\cellcolor{gray!20}72  \\ 
                    & 2-8& 8.34 &96.79 &	99.4 &	8.66 &	88.38 &	95.1 \\ \cline{2-8}
                    & 1-8   & 8.4 &	96.89 &	99.5  &	9.1 &	82.83 &	\cellcolor{gray!20}92.3 \\ \hline
5  			& 1-8  & 7.2 &	93.65 &	97.5 &	\cellcolor{gray!20}6.8 &	88.96 &	\cellcolor{gray!20}93.8 \\   \hline
6  			& 1-8  & \cellcolor{gray!20}13.2 &	66.13 &	85.0 &	8.0 &	72.84 &	85.0 \\   \hline
7  			& 1-8  & \cellcolor{gray!20}10.4 &	62.32 &	83.8 &	7.5 &	67.52 &	83.8 \\   \hline
8  			& 1-8  & \cellcolor{gray!20}12.2 &	93.89 &	98.8 &	9.1 &	88.28 &	93.8 \\   \hline
9  			& 1-8  & \cellcolor{gray!20}22.4 &	86.53 &	96.3 &	8.8 &	88.28 &	96.3 \\   \hline
10  			& 1-8  & \cellcolor{gray!20}14.4 &	93.99 &	98.8 &	8.6 &	93.19 &	96.3 \\   \hline
11  			& 1-8  & \cellcolor{gray!20}11.3 &	96.50 &	100 &	10.1 &	95.06 &	98.8 \\   \hline
12  			& 1-8  & \cellcolor{gray!20}11.2 &	86.20 &	95.0 &	8.0 &	94.11 &	97.5 \\   \hline
13  			& 1-8  & 9.5 &	80.51 &	90.0 &	8.1 &	84.89 &	90.0 \\   \hline\hline
\multicolumn{2}{c|}{\bf{State Avg.}}  & \cellcolor{gray!20}12.4 &	89.03 &	\cellcolor{gray!20}96.1  & \cellcolor{gray!20}8.1 &	87.33 &	\cellcolor{gray!20}93.4 \\
\end{tabular}
\end{center}
\end{table}

\begin{figure}[t]
\begin{center}
\includegraphics[width=\columnwidth]{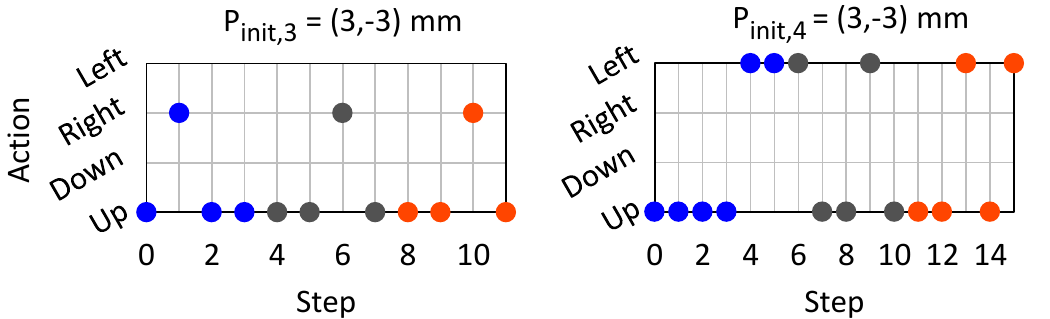}
\caption{Actions for two different initial positions. Hole number: 4; number of episodes: 3 (change of color indicates change of episode)}
\label{fig:actions}
\end{center}
\end{figure}

An example of actions outputted by the DNN are shown in Fig. \ref{fig:actions}. As shown in the figure, the predominant actions for initial position $P_{init,3}$ (bottom left) were to move up and right, and for $P_{init,4}$ (bottom right), to move up and left. These results demonstrate the DNN could correctly identify the actions that lead to the anchor bolt insertion for this two initial positions. For the remaining holes and initial positions, similar results were obtained. 

\begin{figure}[t]
\begin{center}
\includegraphics[width=\columnwidth]{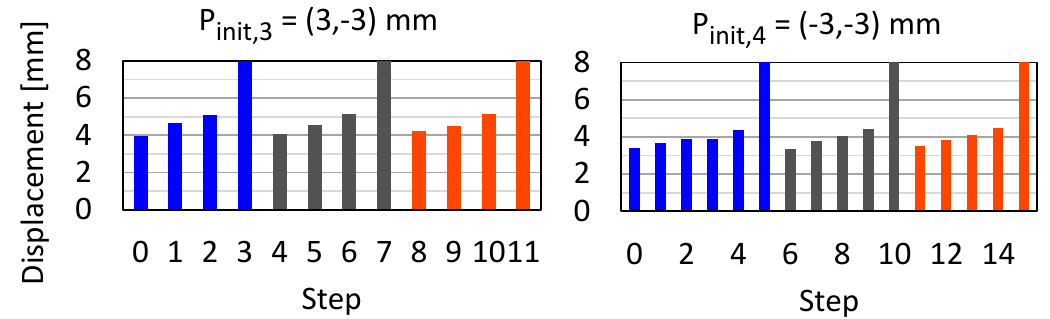}
\caption{Displacement for two different initial positions. Hole number: 4; number of episodes: 3 (change of color indicates change of episode)}
\label{fig:displacement}
\end{center}
\end{figure}

\begin{figure}[t!]
\begin{center}
\includegraphics[width=\columnwidth]{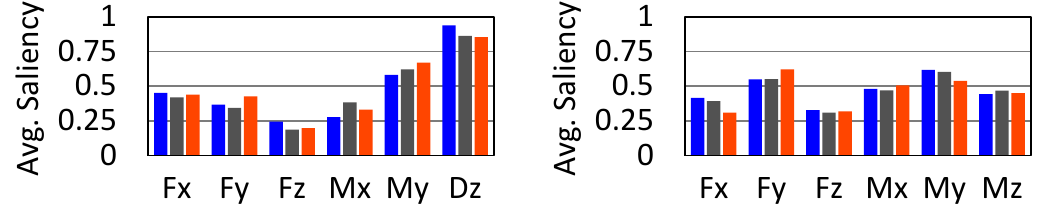}\\[-2ex]
\subfloat[\label{subfig:dispwith} $s_1$: With Dz]{\hspace{.6\columnwidth}}
\subfloat[\label{subfig:dispwithout} $s_2$: Without Dz]{\hspace{0.4\columnwidth}} 
\caption{Average saliency for holes 2 (blue). 3 (gray), and 4 (orange).}
\label{fig:saliency}
\end{center}
\end{figure}

Search failures mainly occurred when the DNN estimated an initial search direction that was away from the hole. In these cases, we assume the DNN encounters a region that does not provide data that enable the prediction of the hole direction, making the peg to cross the search area limits.

\subsection{Displacement importance analysis}\label{subsec:displacementanalysis}

Displacement results obtained during search for three episodes, starting from initial positions $P_{init,3}$ and $P_{init,4}$, are shown in Fig. \ref{fig:displacement}. In all episodes, the robot successfully found the hole. As predicted, the displacement increased as the anchor bolt approached the hole position. This result demonstrates the displacement input presents useful information for accomplishing the hole-search task. 

Average saliency obtained for input states $s_1$ and $s_2$ is shown in Fig. \ref{fig:saliency}. As predicted, average saliency for $D_z$ is higher than that for the other inputs under the same input state (Fig. \ref{subfig:dispwith}). Furthermore, when compared to the saliency of $M_z$ under input state $s_2$, the saliency of $D_z$ is also high, namely, about double the importance. These results confirm that $D_z$ is an important input for determining the correct movement to accomplish hole discovery. Also, they explain, to some extent, the higher success rate when $s_1$ is used.

\subsection{Generalization capability assessment}\label{subsec:resgencapassess}
Results of the generalization capability assessment of the proposed method is shown in Table \ref{tab:resgen}. As shown, even with random initial positions or a different anchor, the proposed method executes the task with high success rate, which indicates the method can generalize well to different conditions. Particularly, for random initial conditions, the results demonstrate the compliance of the gripper can cope with submillimeter misalignment of the peg and hole because step-sizes are 1 mm and initial positions were not exactly 2 or 3 mm away from the hole (e.g., $P_{init}= (2.1, 2.8)$ mm).

\begin{table}[t]
\begin{center} 
\caption{Results of generalization capability assessment}
\label{tab:resgen} 
\setlength\tabcolsep{3.5pt} 
\begin{tabular}{c|ccc|ccc}
\multirow{3}{*}{\bf{Hole}} & \multicolumn{3}{c|}{\bf{Random Init. Pos.}} & \multicolumn{3}{c}{\bf{Pin-type Anchor Bolt}} \\ \cline{2-7}
                           & \multirow{2}{*}{\shortstack{\bf{Avg.}\\\bf{Time [s]}}} & \multirow{2}{*}{\shortstack{\bf{Avg.}\\\bf{Reward}}} & \multirow{2}{*}{\shortstack{\bf{Success}\\\bf{rate [$\%$]}}} & \multirow{2}{*}{\shortstack{\bf{Avg.}\\\bf{Time [s]}}} & \multirow{2}{*}{\shortstack{\bf{Avg.}\\\bf{Reward}}} & \multirow{2}{*}{\shortstack{\bf{Success}\\\bf{rate [$\%$]}}} \\ 
                           &   &  &  &  & \\ \hline
8  &  6.1 &	86.78 &	90.0 & 7.5 & 93.5 & 97.5 \\ 
9  &  7.6 &	88.68 &	94.0 & 7.4 &	79.0 &	 91.3  \\
10 &  7.7 &	92.88 &	98.0 &  9.8 &	80.44 &	90.0 \\ 
11 &  7.7 &	92.88 &	98.0 &  9.0 &	95.25 &	98.8 \\ \hline
Avg.  & 7.3 &	90.30 &	\cellcolor{gray!20}95.0  & 8.4 &	87.03 &	\cellcolor{gray!20}94.4  \\
\end{tabular}
\end{center}
\end{table}

\begin{table}[t]
\begin{center} 
\caption{Evaluation results for model-based methods}
\label{tab:resmodelbased} 
\setlength\tabcolsep{6pt} 
\begin{tabular}{c|ccc|cc}
\multirow{3}{*}{\bf{Hole}} & \multicolumn{3}{c|}{\bf{Blind search}} & \multicolumn{2}{c}{\bf{Moment-based search}} \\ \cline{2-6}
                           & \multirow{2}{*}{\shortstack{\bf{Avg.}\\\bf{Time [s]}}} & \multirow{2}{*}{\shortstack{\bf{Max.}\\\bf{Time [s]}}} & \multirow{2}{*}{\shortstack{\bf{Success}\\\bf{rate [$\%$]}}} & \multirow{2}{*}{\shortstack{\bf{Avg.}\\\bf{Time [s]}}} & \multirow{2}{*}{\shortstack{\bf{Success}\\\bf{rate [$\%$]}}} \\ 
                           &   &  &  &  \\ \hline
2  &  44.5  & \cellcolor{gray!20}102.5  &	100 &	6.3 &	32 \\  
3  &  37.0  & 60.0	&   100 &	14.0 &  16  \\ 
4  &  58.2  & \cellcolor{gray!20}102.5	&   100 &	19.6 &	19  \\ \hline
Avg. & 46.6 &  88.3   &100 &	5.3 &	\cellcolor{gray!20}22 \\
\end{tabular}
\end{center}
\end{table}

\subsection{Comparison with model-based approaches}
The evaluation results for search with the model-based methods are listed in Table \ref{tab:resmodelbased}. As listed, blind search could find the hole 100$\%$ of the times, but the execution time depended on the starting point, taking up to 102.5 seconds to find the hole when starting from $P_{init,2}$. Results of search with the moment-based search showed poor average success rate of 22$\%$, which was mainly caused by the bias of the gripper to tilt to the upper direction independently of the region where the peg contacts the wall border, which was not predicted by the model. The results obtained by these two approaches indicate the search with DNNs present a good trade-off between success rate and execution time.

\section{CONCLUSION}
An off-policy, data-driven method for training an industrial robot to accomplish the peg-in-hole task for holes in concrete was proposed. Results of evaluations of the proposed method show the method enables the robot to find unknown holes with 96.1$\%$ success rate and average execution time of 12.4 seconds. The results also show the proposed method generalizes well to different conditions as it enabled a robot to accomplish the peg-in-hole task with random initial positions and a different type of anchor bolt. Additionally, the results show the DNN used by the proposed method is closer to meeting the requirements of the construction industry, namely perfect success rate and low execution time, since the DNN presented better trade-off between success rate and execution time compared to other two traditional methods. Even though the proposed method was evaluated with anchor bolts with diameter of 12 mm, it is presumed the method can be extended to any cylindrical object with any diameter to be inserted in chamfered holes opened in high-friction materials, which can be easily found in the construction and manufacturing industries.

\addtolength{\textheight}{-10cm}   



\bibliographystyle{IEEEtran}
\bibliography{myBib}

\end{document}